\documentclass{article}
\usepackage{amssymb}

\usepackage[preprint]{corl_2025} 

\usepackage{multirow}
\newcommand{\smbf}[1]{\small\textbf{#1}}

\title{SayCoNav: Utilizing Large Language Models \\ for Adaptive Collaboration \\ in Decentralized Multi-Robot Navigation}

\author{
  Abhinav Rajvanshi\\
  SRI International\\
  \texttt{abhinav.rajvanshi@sri.com} \\
  \And
  Pritish Sahu \\
  SRI International \\
  \texttt{pritish.sahu@sri.com} \\
  \AND
  Tixiao Shan \\
  SRI International \\
  \texttt{tixiao.shan@sri.com} \\
  \And
  Karan Sikka \\
  Meta \\
  \texttt{karansikka@meta.com} \\
  \And
  Han-Pang Chiu \\
  SRI International \\
  \texttt{han-pang.chiu@sri.com} \\
}

\begin{document}
\maketitle


\begin{abstract}
    Adaptive collaboration is critical to a team of autonomous robots to perform complicated navigation tasks in large-scale unknown environments. An effective collaboration strategy should be determined and adapted according to each robot’s skills and current status to successfully achieve the shared goal. We present SayCoNav, a new approach that leverages large language models (LLMs) for automatically generating this collaboration strategy among a team of robots. Building on the collaboration strategy, each robot uses the LLM to generate its plans and actions in a decentralized way. By sharing information to each other during navigation, each robot also continuously updates its step-by-step plans accordingly. We evaluate SayCoNav on Multi-Object Navigation (MultiON) tasks, that require the team of the robots to utilize their complementary strengths to efficiently search multiple different objects in unknown environments. By validating SayCoNav with varied team compositions and conditions against baseline methods, our experimental results show that SayCoNav can improve search efficiency by at most 44.28\% through effective collaboration among heterogeneous robots. It can also dynamically adapt to the changing conditions during task execution. 
\end{abstract}



\section{Introduction}
	Enabling a team of heterogeneous robots to collaborate for conducting a complicated navigation task in a new large-scale environment is critical to many applications, such as searching multiple objects in a big house. The best collaboration approach shall depend on the team composition. Robots shall leverage the diverse skills from each other, leading to better outcomes under a joint goal. In addition, the collaborative strategy may need to be adjusted during the mission, in response to changing conditions from any autonomous robot in the team. 

    However, most multi-robot navigation systems rely on either centralized architectures \cite{RobotSurvey} or manually-defined distributed processes \cite{DistributedSurvey} for collaboration and communication. These systems lack the flexibility and efficiency for handling the complexities of navigation tasks in the real word. Ideally, as humans do, the multi-robot system should be decentralized and adaptive. Each robot reasons its perceived scenes and dynamically shares information to other robots during exploration. Based on the gathered information among the team, each robot continuously updates its own plans and adjusts the collaboration strategy for achieving the joint mission goal. 

    Recent advancements in Large Language Models (LLMs) offer an attractive opportunity to promote multi-robot navigation systems to decentralized planning and dynamic collaboration. LLMs have demonstrated promising capabilities in robotic planning, human knowledge utilization, and dialogue-based communication. However, almost all LLM-based navigation works still focus on applications to using either a single robot \cite{SayNav, SayPlan} or a group of homogeneous robots \cite{Co-NavGPT, MCoCoNav}. Leveraging LLMs for decentralized navigation using heterogeneous robots is still an unexplored area.  

    In this paper, we propose SayCoNav, a new approach that leverages LLMs for enabling adaptive collaboration among multiple autonomous robots for decentralized planning and navigation. The key innovation of SayCoNav is to utilize LLMs to automatically generate the collaboration strategy based on the skills of each robot in the team. Specifically, SayCoNav enables each robot to share its background information, such as its capability specification and operational constraints, to other robots prior to the mission. One randomly selected robot then feeds the assigned team task with the team background information into the LLM, for retrieving appropriate knowledge to form the collaboration strategy. This mechanism guarantees that the LLM generates feasible collaboration strategies that leverage the strengths of different robots to solve complex navigation tasks. When any robot has changes in its physical condition during navigation, it can ground the LLM again for generating a new collaboration strategy which accommodates its changed condition. The updated strategy will then be shared and executed among the team. This way ensures that the team adapts its strategies and interactions to effectively work together under dynamic situations.

\begin{figure*} [t]
  \centering
   \includegraphics[width=1\linewidth]{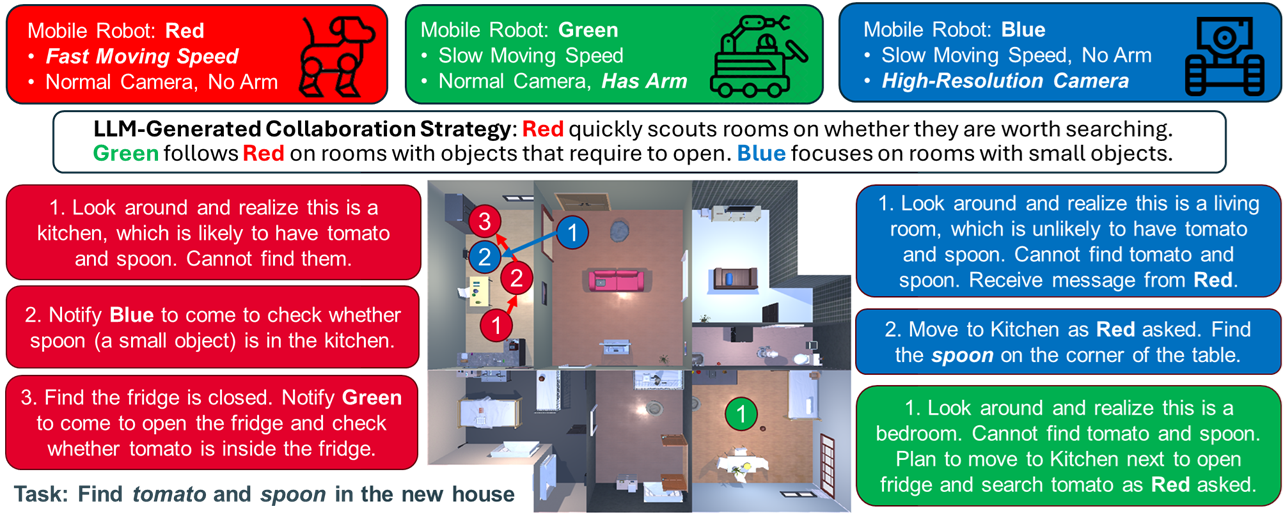}
   \caption{A SayCoNav example: Three heterogeneous robots use LLMs to define and to execute the collaboration strategy based on their skills, for searching two target objects in a new house. }
   \label{fig:concept}
\end{figure*}

    SayCoNav utilizes a novel three-level decentralized planning architecture to handle the complexity from cooperative navigation tasks in large-scale new environments. Each robot uses a top-level LLM-based global planner to share information among the team for defining the collaboration strategy. A middle-level local planner leverages the LLM to dynamically generate step-by-step plans for the individual robot based on the collaboration strategy. The step-by-step plan is then executed by a bottom-level action planner, that maps each step into a sequence of primitive actions. Figure \ref{fig:concept} illustrates an example of SayCoNav  using this decentralized planning architecture among three robots, for searching a tomato and a spoon in the new house. Based on the LLM-generated collaboration strategy, the Red robot utilizes its fast speed to quickly scout the rooms and finds that the kitchen is likely to have the two target objects. The Red robot then calls the Blue robot to come to use its high-resolution camera for locating the small spoon. It also requests the Green robot, which has the manipulation arm, to open the fridge and to check whether the tomato is inside the fridge.
    
    To fully validate SayCoNav, we create a benchmark dataset via the ProcTHOR framework \cite{ProcTHOR} for a complex navigation task, Multi-Object Navigation (MultiON) \cite{MultiON}. We configure a large amount of episodes from this benchmark dataset, with a wide variety of room layouts, furniture arrangements, and object choices. The design of these episodes requires the robot team to leverage skills from each other to efficiently search multiple different objects in an unknown large-scale house. Our experiments demonstrate that SayCoNav can effectively enable robots collaborate with each other, even for these long horizon tasks in unknown environments. The results also show that SayCoNav can generate customized strategies across different team compositions and conditions.  
  

\section{Related Work}
\label{sec:related}

\subsection{Using LLMs for Single-Robot Navigation}

Planning with LLMs has become an emergent trend to single-robot navigation. Compared to learning-based methods \cite{DDPPO} which generally require at least hundreds of millions of iterations for training an agent (i.e. a simulated robot) to navigate in new environments, LLMs have demonstrated excellent capabilities to perform a new navigation task \cite{SayNav, SayPlan, L3MVN, OrionNav, UniGoal, SGNav, Graph2Nav} with only a few training examples via in-context learning \cite{FewShotLLM, InContextLearning}. These works typically use a two-level planning architecture. A high-level planner utilizes either 3D scene graphs \cite{SayNav, SayPlan, OrionNav, UniGoal, SGNav, Graph2Nav} or semantic maps \cite{L3MVN} of the explored environment to ground LLMs for generating step-by-step plans. Each planned step is then executed by an oracle (ground truth) planner or a pre-trained low-level planner, that maps one step into a sequence of primitive actions for execution.

Inspired by these works on single-robot navigation, SayCoNav utilizes a novel LLM-based three-level decentralized planning architecture for multi-robot navigation. The top-level global planner is specifically designed for multi-robot coordination, which is beyond the capabilities of single-robot navigation systems. The functionalities of the middle-level local planner from SayCoNav are similar to the high-level planner in single-navigation systems. However, in addition to utilizing LLMs to generate the search plan inside a single room, SayCoNav also leverages LLMs to automatically construct the transition plan (such as which door or room shall go next) based on the collaboration strategy. In contrast, the high-level planner from single-robot navigation systems (such as \cite{SayNav}) typically uses manually-defined heuristic rules for transition across rooms.    

\subsection{Multi-Robot Navigation}

There is a long history on using learning-based methods, such as multi-agent reinforcement learning (MARL) \cite{MultiAgentDRLExplore,AsynMultiAgentDRLExplore}, for multi-robot navigation. Due to the trend in utilizing LLMs for single-robot navigation, there are recent works in leveraging the generalization capabilities from LLMs for multi-robot navigation. However, almost all these works focus on navigation tasks using a team of homogeneous robots \cite{Co-NavGPT,MCoCoNav,CAMON,MHRC}. They either use a centralized LLM-based planner to distribute the task among the team \cite{Co-NavGPT} or manually define the coordination process across decentralized robots \cite{MCoCoNav,CAMON}, since the collaboration strategy among robots with identical skills is straightforward.    

The closest work to SayCoNav is MHRC \cite{MHRC}, which also touches the decentralized planning problem among heterogeneous robots. MHRC only shows one specific team composition (a mobile navigator, a mobile manipulator, and a static manipulator) in their experiments. It decomposes a task into multiple sub-tasks, and manually assigns a role with sub-tasks to each robot based on its skills prior to the mission. The communication process among the robots is also manually designed based on the team composition (the mobile manipulator plays the coordinator role). Note, these configurations cannot be changed during the mission. Each robot in MHRC then uses a two-level planning architecture (similar to \cite{SayNav, SayPlan}), which utilizes 3D scene graphs to ground the LLM for generating step-by-step plans, based on communication messages among the team. 

In contrast, SayCoNav utilizes a novel three-level decentralized planning architecture for automatic generation and execution of the collaboration strategies among different team compositions. Each robot uses the LLM to share background information to other robots. A randomly-chosen agent then uses the LLM to generate the entire collaboration strategy, including communication process among the team and roles to each robot, based on the team background information. The cooperation mechanism can also be updated, if the physical condition of any robot is changed during the mission. 

To the best of our knowledge, SayCoNav provides the first LLM-based adaptive collaboration framework among heterogeneous robots for navigation tasks. The LLM-generated strategy leverages different skills from robots to efficiently explore the unknown environments and search the target objects. The strategy can be automatically generated based on different team compositions and conditions, which is crucial to handle the dynamic and complicated situations in the real world. 


\section{SayCoNav}
\label{sec:SayCoNav}

\begin{figure*} [t]
  \centering
   \includegraphics[width=1\linewidth]{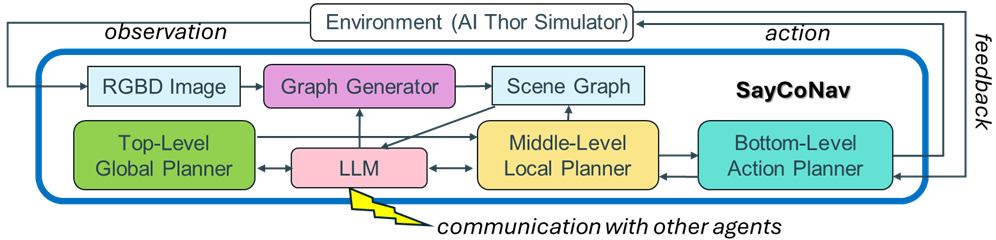}
   \caption{The SayCoNav framework for a single robot.}
   \label{fig:framework}
\end{figure*}


SayCoNav is designed to enable adaptive collaboration capabilities in a decentralized team of heterogeneous robots, for performing complicated navigation tasks in large-scale new environments. To validate SayCoNav, we choose Multi-Object Navigation task (MultiON) \cite{MultiON}, which has much higher planning complexities than typical object-goal navigation tasks that look for only a single object \cite{ObjectGoal}. Our goal of MultiON is to navigate a team of the agents (simulated robots) in a large-scale unknown environment in order to find an instance for each of multiple predefined object categories. Each agent is initialized at a random location in the environment. At each time instance during navigation, each agent receives environment observations and takes actions. The task execution is successful if the team of robots locates (by detection) all three objects within a time period.

Note, the observation quality and the action space for each agent depends on its skills and conditions, as shown in Figure \ref{fig:concept}. For example, a robot with a high-resolution camera perceives better RGBD images and semantic segmentation maps. The mobile robot with manipulator arm can issue manipulation control commands (\textit{open} and \textit{close}) in addition to movement control commands (\textit{turn-left}, \textit{turn-right}, \textit{move-forward}, \textit{stop}, and \textit{look-around}).  Note, we specifically chose many different object categories which are either placed inside closed entities (such as a tomato in a fridge) or difficult to be observed due to their small sizes (such as a spoon). The team of the robots is required to dynamically coordinate the search plan, depending on the perceived surroundings (i.e. the kitchen may have the tomato and spoon) and different skills (i.e. high-resolution cameras can better locate the spoon) across the agents, based on a collaboration strategy. 

Figure \ref{fig:framework} illustrates SayCoNav’s framework for each robot. It includes four modules: (1) Top-Level Global Planner, (2) Middle-Level Local Planner, (3) Graph Generator, and (4) Bottom-Level Action Planner. Note, both Top-Level Global Planner and Middle-Level Local Planner are LLM-based planners, that enable adaptive collaboration and decentralized planning for multi-robot navigation. 

\subsection{Top-Level Global Planner}
\label{sec:globalplanner}

\begin{figure*}
  \centering
   \includegraphics[width=1\linewidth]{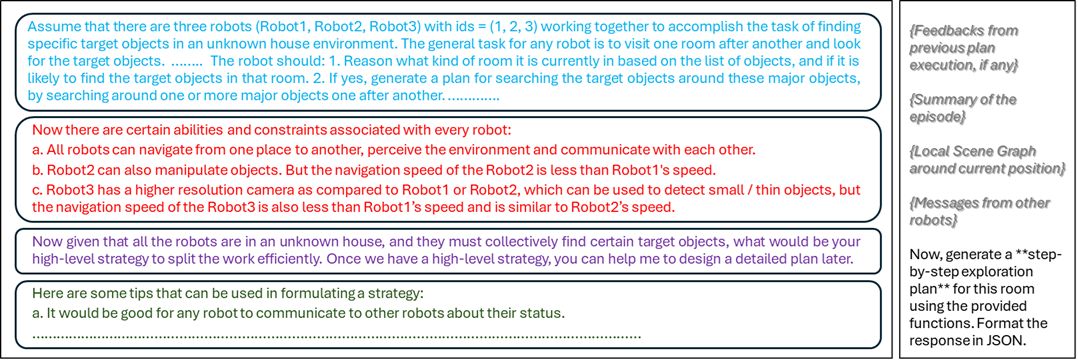}
   \caption{Prompts for top-level global planner (left) and middle-level local planner (right). The global planner prompt includes team task statement (blue), robot skills and conditions (red), collaboration strategy request (purple), and tips (green). The local planner prompt has four components prior to the request to LLM: feedback, summary, local scene graph, and communication messages.}
   \label{fig:prompt}
\end{figure*}

The Top-Level Global Planner uses the LLM to communicate background information among the team for defining the collaboration strategy. Figure \ref{fig:prompt} (left) shows the system prompt for this process. The prompt includes four components: team task statement, robot skills and conditions, collaboration strategy request, and tips. The component of robot skills and conditions is automatically generated based on communication across robots, while other three components are manually specified by the human user based on the MultiON task. We use the chain-of-thought (CoT) prompting technique \cite{CoT} in two components (team task statement and tips). This technique guides the LLM to reason step-by-step as humans do, by decomposing MultiON into logical reasoning steps and describing these steps in the prompt. Due to page limitation, we omit some CoT details in Figure \ref{fig:prompt}. 

The global planner from a randomly chosen agent then feeds this prompt to the LLM, for generating the collaboration strategy prior to the mission. The strategy includes the role assigned to each agent, the communication mechanism among agents, and the task distribution among the team. By grounding the LLM using the specifications of robot skills within the prompt, the generated strategy shall leverage the strengths of different agents to solve the navigation task. The strategy is then shared among the team for execution. The global planner also monitors the condition of the agent. When the agent has condition changes during the mission, it can generate the strategy again.

\subsection{Middle-Level Local Planner}
\label{sec:localplanner}

The Middle-Level Local Planner uses the LLM to dynamically generate step-by-step plans for the individual agent based on the collaboration strategy from the Top-Level Global Planner. Note, each agent is provided with a list of high-level behavior functions, such as \textit{NavigationTo} and \textit{OpenObject}, based on its individual skills. Therefore, the LLM-generated step-by-step plan can be viewed as the composition of these behavior functions from the agent. 

Specifically, each agent automatically generates and updates a short-term plan regularly using the dynamic prompt in Figure \ref{fig:prompt} (right) during the mission. The prompt includes four dynamic components that evolve over time: feedback, summary, local scene graph, and communication messages. The feedback component describes the response from previous plan execution, such as whether it was successful. The summary component illustrates the task status and progress (such as remaining target objects and rooms already visited), automatically generated by the execution of behavior functions. It serves as a memory mechanism, which improves LLM's capabilities to handle long horizon tasks. The component of local scene graph is the abstracted representation of current perceived scenes. The communication messages include information received from other agents. These four components in the prompt ensure that the plan is continuously refined based on newest information among the team, for efficiently achieving the joint goal in a decentralized way. 

Note, the spirit of SayCoNav's local planner is similar to the high-level planner in LLM-based single-robot navigation systems \cite{SayNav, SayPlan}. However, the high-level planner in these systems typically combines heuristic rules (such as which rooms to go next) and LLM-enabled planning (such as the priorities of potential locations inside a room for searching objects). In contrast, SayCoNav's local planner fully relies on the LLM to generate both search plans within rooms and transition plans across rooms, based on the collaboration strategy defined from the global planner. 

\subsection{Graph Generator and Bottom-Level Action Planner}
\label{sec:actionplanner}

Both the Graph Generator and the Bottom-Level Action Planner in SayCoNav are extended from what used in \cite{SayNav}. The graph generator continuously builds and expands a local 3D scene graph of the perceived environment being explored by the agent, based on its received RGBD images. The LLM also retrieves information from communication messages, such as which rooms have been explored by other agents, to augment the local scene graph. The middle-level local planner continuously extracts a subgraph from the full local 3D scene graph based on the current agent position, and converts it into text in the dynamic prompt (Figure \ref{fig:prompt} (right)) to the LLM.

The Bottom-Level Action Planner, which can be either an oracle (ground truth) planner or a pre-trained planner, maps each planned step from the local planner into a sequence of primitive actions. For example, the \textit{NavigationTo} step (a short-distance point-goal navigation sub-task) can be executed by a combination of control commands (\textit{turn-left}, \textit{turn-right}, and \textit{move-forward}). Note for our experiments, we use A* based oracle planner for navigation (same as in \cite{SayNav}) and oracle actions for manipulation in SayCoNav and baseline methods. This way we can focus on evaluating the impact from the global planner and the local planner to multi-robot navigation tasks.


\section{Experimental Results}
\label{sec:result}

We used the ProcTHOR framework \cite{ProcTHOR} with the AI2-THOR simulator \cite{AI2Thor} to conduct MultiON for our experiments. Note, we chose ProcTHOR framework due to its capabilities to generate complicated and realistic large-scale environments for MultiON. Given a house specification (such as 3 bedrooms, 2 bathrooms, 1 living room, and 1 kitchen), it can generate interactive scenes from procedurally-generated full floor plans of a house using 108 object types (including furniture and household entities) with realistic and natural placements. We built a benchmark dataset of 50 episodes using this framework. Each episode is a different house (3-10 rooms) and we select 3 different objects (such as ``laptop”, ``tomato”, and ``spoon”) in each house as the target objects. However, the AI2-THOR simulator with the ProcTHOR framework is originally designed for single-agent simulation. Therefore, we also implemented a mechanism on top of it to enable multi-agent simulation for our experiments.

We assume that each agent can have diverse skills based on its sensors and equipment. We design three special skills (as shown in Figure \ref{fig:concept}): Fast Movement (F), Manipulation (M), and High-Resolution Cameras (H). Without these special skills, an agent is configured as normal speed, no manipulation arm, and with normal RGBD cameras. To validate whether SayCoNav can leverage different skills across robots, the three target objects in each episode are configured to include one normal household object, one object inside a closed entity (such as a tomato inside a fridge) and one small-size object (such as a spoon, a credit card, or a pen).  

We implemented SayNav \cite{SayNav}, a LLM-based single-robot navigation system, as the baseline method. Note, SayNav is originally designed for robots without manipulation arms. Therefore, we extended SayNav with the oracle-based actions to enable manipulation. We conducted experiments to compare SayCoNav with this SayNav-based method across different team compositions and conditions. We use gpt-4o-mini \cite{gpt-4o-mini} as the LLM to each agent for our experiments. We also conducted ablation studies to verify the impact of different implementation choices from SayCoNav. All the simulation experiments are conducted using Nvidia GeForce RTX 2080 Ti GPU.   

\begin{table}
    \centering
    \caption{The performance of SayCoNav and baseline methods on MultiON task. $A_1$, $A_2$ \& $A_3$ denote three different agents, with special skills (F:Fast, M:Manipulation, H:High Res. Camera).}
    \begin{tabular}{c|c|c|ccccc}
        \multirow{2}{*}{\smbf{Agent Type}} & \multirow{2}{*}{\smbf{\#A}} & \multirow{2}{*}{\smbf{Agent Skills}} & \multirow{2}{*}{\smbf{SR(\%)}} & \multirow{2}{*}{\begin{tabular}[c]{@{}c@{}} \smbf{Avg Ep} \\ \smbf{Time (s)}\end{tabular}} & \multicolumn{3}{c}{\smbf{Avg Steps}}\\
        
        & & & & & \small \boldmath$A_1$ & \small \boldmath$A_2$ & \small \boldmath$A_3$ \\
        \hline
        \small SayNav (Homo) & \small 1 & \small $A_1$(F, M, H) & \small 96.0 & \small 363.92 & \small 508.85 & -  & - \\
        \hline
        \small SayNav (Homo) & \small 1 & \small $A_1$(M, H) & \small 96.0 & \small 435.15 & \small 508.85  & - & -\\
        \hline
        \small SayNav (Homo) & \multirow{2}{*}{\small 3} & \multirow{2}{*}{\small $All$ (F, M, H)} & \small 96.0 & \small 252.17 & \small 353.52 & \small 351.81 & \small 377.83  \\
        \small \textbf{SayCoNav} (Homo) &  &  & \small 96.0 & \small 195.98 & \small 208.83 & \small 202.94 & \small 218.81  \\
        \hline
        \hline
        \small SayNav (Hetero) & \multirow{2}{*}{\small 2} & \multirow{2}{*}{\small $A_1$(F, H), $A_2$(M)} & \small 88.0 & \small 441.36 & \small 764.25 & \small 424.50 & - \\
        \small \textbf{SayCoNav} (Hetero) &  &  & \small 92.0 & \small 267.46 & \small 386.76 & \small 285.85 & - \\
        \hline
        \small SayNav (Hetero) & \multirow{2}{*}{\small 2} & \multirow{2}{*}{\small $A_1$(H), $A_2$(F, M)} & \small 90.0 & \small 546.32 & \small 627.89  & \small 435.09 & - \\
        \small \textbf{SayCoNav} (Hetero) &  &  & \small 90.0 & \small 256.17 & \small 286.87  & \small 305.51 & - \\
        
        \hline
        \small SayNav (Hetero) & \multirow{2}{*}{\small 3} & \multirow{2}{*}{\small $A_1$(F), $A_2$(M), $A_3$(H)} & \small 90.0 & \small 537.43 & \small 515.82 & \small 456.76 & \small 707.98  \\
        \small \textbf{SayCoNav} (Hetero) &  &  & \small 90.0 & \small 299.45 & \small 393.11 & \small 287.67 & \small 336.02  \\
    \end{tabular}
    \label{tbl:res_main}
\end{table}

\subsection{Homogeneous Agents}

Table \ref{tbl:res_main} summarizes the evaluation of SayCoNav against the baseline method over different team compositions. Note for each experimental setting (a row in the table) over 50 episodes, we report successful rate (SR, \%), average episode time (seconds), and average number of navigation steps for each agent in the team. Note, we do not use SPL metrics for evaluation, because it is difficult to identify the ground truth paths for multiple distributed agents. In addition, the average episode time and the average number of navigation steps are computed from successful episodes (i.e. failure episodes take much more time and steps).  

The top portion of Table \ref{tbl:res_main} shows the results using homogeneous agents. Note, the SayNav-based baseline method can complete 96\% of episodes using purely one agent, while the fast movement skill shortens the navigation time (from 435.15 to 363.92 seconds) using same number of steps. 

When the team includes three agents with identical skills, both SayNav and SayCoNav can complete scenarios faster than using only one agent. However, SayCoNav's LLM-based collaboration strategy is to uniformly distribute the search task among the team by continuously sharing the progress to each other. Compared to SayNav, which navigates three agents independently without coordination, SayCoNav efficiently reduces the search time (from 252.17 to 195.98 seconds). The similar number of steps from each agent indicates the successful distribution of the task among the team.   

\subsection{Heterogeneous Agents}

The bottom portion of Table \ref{tbl:res_main} shows the results using heterogeneous agents. We tested two cases using two agents with different skill sets. For each case, SayCoNav is able to automatically generate a customized collaboration strategy to leverage skills from two agents. Compared to SayNav, the strategy significantly reduces the search time by utilizing the fast speed skill (more navigation steps) for the correspondent agent ($A_{1}$ in case 1, $A_{2}$  in case 2) to scout rooms. 

For three heterogeneous agents, SayCoNav automatically defines the strategy (as shown in Figure \ref{fig:concept}) to leverage the special skill ($A_{1}(F), A_{2}(M), A_{3}(H)$) from each agent. Compared to SayNav, the search time is reduced by $44.28\%$, indicating the influence from SayCoNav's collaboration strategy.

\begin{table}
    \centering
    \caption{Impact of adaptive collaboration within episodes. Based on definition in Table \ref{tbl:res_main}, the initial skill for agents is $A_{1}(F)$, $A_{2}(M)$, and $A_{3}(H)$. The condition of $A_{1}$ changes within each episode.}
    \begin{tabular}{c|ccccc}
        \multirow{2}{*}{\smbf{Experiment}} & \multirow{2}{*}{\smbf{SR(\%)}} & \multirow{2}{*}{\begin{tabular}[c]{@{}c@{}} \smbf{Avg Ep} \\ \smbf{Time (s)}\end{tabular}} & \multicolumn{3}{c}{\smbf{Avg Steps}}\\
        
        & & & \small \boldmath$A_1$ & \small \boldmath$A_2$ & \small \boldmath$A_3$ \\
        \hline
        \small SayCoNav (w/o adaptive strategy) & \small 88.0 & \small 331.37 & \small 309.23 & 301.18  & 332.30 \\
        \small SayCoNav (w. adaptive strategy)  & \small 92.0 & \small 319.31 & \small 255.17  & 323.52 & 307.54 \\
        
    \end{tabular}
    \label{tbl:res_adapt}
\end{table}

\subsection{Impact of Adaptive Collaboration during Task Execution}

Table \ref{tbl:res_adapt} shows the performance of SayCoNav in response to changes in the physical condition of the agent. For these experiments, we set up three heterogeneous agents with different special skills ($A_{1}(F), A_{2}(M), A_{3}(H)$) as initial conditions. Then for each of the 50 episodes, we configured the fast-speed agent $A_1$ to run out of battery during navigation and then move slower than other agents.  

For this physical condition change in $A_1$, the global planner in $A_1$ can automatically ground the LLM again using the prompt (Figure \ref{fig:prompt} (left)) with an updated component of robot skill and condition. It then generates a new collaboration strategy to accommodate the condition change from $A_1$. Specifically, the new strategy moves more responsibilities (such as scouting) from $A_1$ to $A_2$. The reason is that $A_3$ requires more time to conduct detailed search by using its high-resolution camera to search small objects inside a new room. In comparison, $A_2$ is more suitable to play more roles even it has same moving speed as $A_3$. The new strategy also assigns $A_1$ to assist $A_3$ when needed. 

By adapting the collaboration strategy within the episode, SayCoNav is able to increase the successful rate (from 88\% to 92\%) and to reduce the search time (from 331.37 to 319.31 seconds), compared to using the same strategy from beginning to the end. From Table \ref{tbl:res_adapt}, with adaptive strategy, it is clear that the number of steps taken by $A_1$ becomes lower while $A_2$ takes more steps. It indicates the influence from adaptive collaboration in response to the condition change of $A_1$.

\begin{table}
    \centering
    \caption{The performance of SayCoNav using collaboration strategies from LLMs and humans. Based on definition in Table \ref{tbl:res_main}, the skill set for agents is $A_{1}(F)$, $A_{2}(M)$, and $A_{3}(H)$. }
    \begin{tabular}{c|c|ccccc}
        \multirow{2}{*}{\smbf{Agent Type}} & \multirow{2}{*}{\smbf{Collaboration Strategy}} & \multirow{2}{*}{\smbf{SR(\%)}} & \multirow{2}{*}{\begin{tabular}[c]{@{}c@{}} \smbf{Avg Ep} \\ \smbf{Time (s)}\end{tabular}} & \multicolumn{3}{c}{\smbf{Avg Steps}}\\
        
        & & & & \small \boldmath$A_1$ & \small \boldmath$A_2$ & \small \boldmath$A_3$ \\
        \hline        
        \small SayCoNav (Hetero) & LLM-Generated & \small 90.0 & \small 299.45 & \small 393.11 & \small 287.67 & \small 336.02  \\
        \small SayCoNav (Hetero)  & Human-Refined & \small 90.0 & \small 309.90  & \small 383.40 & \small 303.62 & \small 296.31  \\
    \end{tabular}
    \label{tbl:res_human}
\end{table}

\subsection{Difference between Strategies from LLMs and Humans}

To verify whether the LLM-generated collaboration strategy is ideal, we also asked a human expert to review and revise the strategy generated from the global planner. We configured three heterogeneous agents with different special skills ($A_{1}(F), A_{2}(M), A_{3}(H)$). The human expert augments the LLM's collaboration strategy (Figure \ref{fig:concept}), by explicitly requesting each agent to avoid searching rooms which are currently being explored by other robots unless there is no room left to explore. The expert also recommends the agent to remember the recently used doors and look to explore other doors for transition across rooms. Table \ref{tbl:res_human} shows the human-refined strategy achieves similar performance, which validates the feasibility and effectiveness of LLM-generated strategies.

\subsection{Additional Ablation Study}

We also conducted additional ablation studies for different implementation choices in SayCoNav, with results using three heterogeneous agents in Table \ref{tbl:res_ablation}. The first set of experiments verifies the contribution of different components to the dynamic prompt in the local planner (Figure \ref{fig:prompt} (right)). It is clear that the performance becomes worse, without either feedback or summary in the prompt. 

The second set of experiments uses different LLMs (gpt-4o-mini \cite{gpt-4o-mini} and o4-mini \cite{o4-mini}) for SayCoNav. We focus on evaluating small-size LLMs, due to our plan on incorporating LLMs into the processor with real robots in the future. The performance is slightly improved using o4-mini, which is a reasoning model designed specifically for planning tasks. These results demonstrate that SayCoNav can leverage different LLMs for collaborative navigation applications.

\begin{table}
    \centering
    \caption{Additional ablation studies using three heterogeneous agents ($A_{1}(F)$, $A_{2}(M)$, and $A_{3}(H)$), based on the definition in Table \ref{tbl:res_main}.}
    \begin{tabular}{c|c|ccccc}
        \multirow{2}{*}{\smbf{Experiment}} & \multirow{2}{*}{\smbf{LLM}} & \smbf{SR} & \multirow{2}{*}{\begin{tabular}[c]{@{}c@{}} \smbf{Avg Ep} \\ \smbf{Time (s)}\end{tabular}} & \multicolumn{3}{c}{\smbf{Avg Steps}}\\
        
        & & (\%) & & \small \boldmath$A_1$ & \small \boldmath$A_2$ & \small \boldmath$A_3$ \\
        \hline
               \small SayCoNav & \small gpt-4o-mini  & \small 90.0 & \small 299.45 & \small 393.11 & \small 287.67 & \small 336.02  \\
        \hline
        \small SayCoNav (w/o feedback) & \small gpt-4o-mini & \small 84.0 & \small 325.90 & \small 407.10 & \small 323.05  & \small 315.98 \\
        \small SayCoNav (w/o feedback \& summary) & \small gpt-4o-mini  & \small 78.0 & \small 321.60 & \small 446.46  & \small 323.59 & \small 308.44 \\
        \hline
 
        \small SayCoNav & \small o4-mini  & \small 92.0 & \small 293.87 & \small 351.78 & \small 255.43 & \small 301.50  \\
    \end{tabular}
    \label{tbl:res_ablation}
\end{table}


\section{Conclusion}
\label{sec:conclusion}

We present SayCoNav, a new approach that utilizes LLMs for generating and executing collaboration strategies among a decentralized team of heterogeneous robots, to efficiently search multiple different objects in unknown environments. SayCoNav enables adaptive collaboration capabilities, by automatically designing the collaboration strategy that leverages diverse skills among the team and accommodates the physical condition changes from any robot. We validate SayCoNav with varied
team compositions and conditions against baseline methods. Our experimental results and ablation studies demonstrate the effectiveness and efficiency of SayCoNav for complicated multi-robot navigation tasks in large-scale unknown environments.


\section{Limitation}
\label{sec:limitation}

\begin{figure*}
  \centering
   \includegraphics[width=1\linewidth]{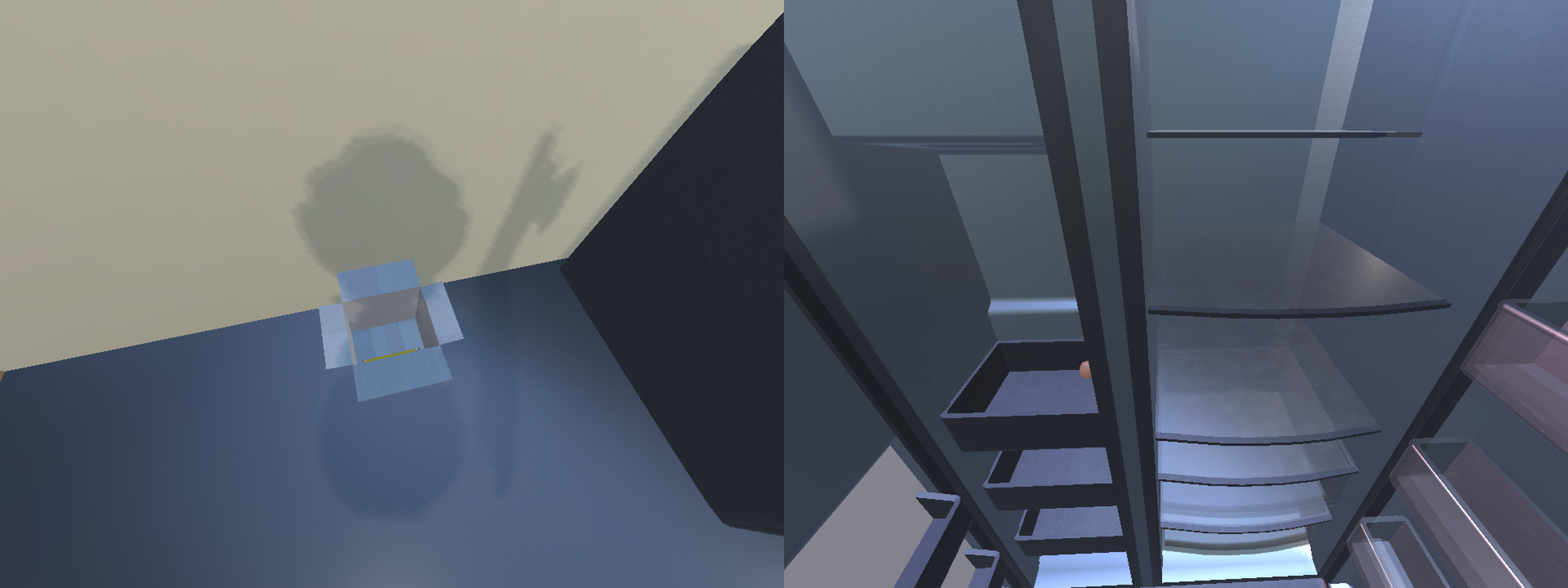}
   \caption{Example of failure cases where a small pencil is present inside a box (left) and a small egg is inside a fridge (right). }
   \label{fig:failure}
\end{figure*}

In this section, we would like to discuss two limitations of SayCoNav that we discovered from the failure episodes in Section \ref{sec:result}. First, SayCoNav leverages LLMs to generate strategies for efficient collaborations among a team of robots. Yet, we encountered certain scenarios in our experiments which would demand very complex collaboration strategies beyond what LLMs generate to successfully complete the task. For example, there were cases which required opening of an object to find a tiny object inside it, such as a pencil inside a closed box or a small egg deep inside a fridge, as shown in Figure \ref{fig:failure}. Detection of such target objects requires both manipulation arms and high-resolution cameras. It's an easy task for a single agent with both these capabilities. However, for a team of agents in Figure \ref{fig:concept}, it requires the green robot (has arm) and the blue robot (with high resolution camera) to work closely together to achieve this task. This was something we found that LLM couldn't handle it well and was majorly the reason of failure cases in our experiments.

Second, we employed various mechanisms in minimizing LLM hallucinations, such as providing feedback and summary at every stage (Figure \ref{fig:prompt}) and making LLMs generate an explanation for each step.  However, we still came across a few cases with common LLM hallucination problems. These examples include ignoring an unexplored door and doing redundant searches. We provide more details about our algorithms, including how we deal with LLM hallucinations, and experiments in the supplementary material. 

As a part of future work, we would like to port SayCoNav to a real multi-robot system. LLM-based single-agent systems, such as \cite{SayNav, SayPlan, OrionNav, Graph2Nav}, have been successfully ported to a real robot for operation. Since our graph generator and bottom level planner were extended from \cite{SayNav}, we expect that the transition of SayCoNav to real robots can be natural and straightforward.


\clearpage


\bibliography{corl2025}  

\end{document}